\documentclass[letterpaper, 10 pt, conference]{ieeeconf}

\usepackage{amsmath}
\usepackage{array}
\usepackage{graphicx}
\usepackage{amssymb}
\usepackage{multirow}
\usepackage{comment}
\usepackage{tabularx,booktabs,makecell,threeparttable}
\newcolumntype{C}{>{\centering\arraybackslash}X}

\usepackage{cite}
\usepackage{float}

\usepackage{subcaption}

\newcommand{\appropto}{\mathrel{\vcenter{
  \offinterlineskip\halign{\hfil$##$\cr
    \propto\cr\noalign{\kern2pt}\sim\cr\noalign{\kern-2pt}}}}}
    
\IEEEoverridecommandlockouts                              
\usepackage[dvipsnames]{xcolor}

\title{\LARGE \bf
Proficiency Constrained Multi-Agent Reinforcement Learning for Environment-Adaptive Multi UAV-UGV Teaming}

\author{Qifei Yu$^{+}$, Zhexin Shen$^{+}$, Yijiang Pang$^{+}$, Rui Liu$^{*}$

\thanks {Authors are with Cognitive Robotics and AI Lab (CRAI), College of Aeronautics and Engineering, Kent State University, Kent, OH 44240, USA}%
\thanks{$^{*}$Rui Liu is the corresponding author ruiliu.robotics@gmail.com. $^{+}$The first three authors have equal contributions to this paper. }%
}

\begin{document}

\maketitle
\thispagestyle{empty}
\pagestyle{empty}

\begin{abstract} 
A mixed aerial and ground robot team, which includes both unmanned ground vehicles (UGVs) and unmanned aerial vehicles (UAVs), is widely used for disaster rescue, social security, precision agriculture, and military missions. However, team capability and corresponding configuration vary since robots have different motion speeds, perceiving ranges, reaching areas, and resilient capabilities to the dynamic environment. 
Due to heterogeneous robots inside a team and the resilient capabilities of robots, it is challenging to perform a task with an optimal balance between reasonable task allocations and maximum utilization of robot capability. To address this challenge for effective mixed ground and aerial teaming, this paper developed a novel teaming method, proficiency aware multi-agent deep reinforcement learning (Mix-RL), to guide ground and aerial cooperation by considering the best alignments between robot capabilities, task requirements, and environment conditions. Mix-RL largely exploits robot capabilities while being aware of the adaption of robot capabilities to task requirements and environment conditions. 
Mix-RL's effectiveness in guiding mixed teaming was validated with the task "social security for criminal vehicle tracking".

\end{abstract}


\section{INTRODUCTION}
Diversity of robot function and team size enables a mixed aerial-ground robot system, including both Unmanned Aerial Vehicle (UAV) and Unmanned Ground Vehicle (UGV), to perform complex tasks with large area coverage and environment dynamics. 
In the context of natural disaster search-and-aid tasks \cite{resilient_disaster_site_exploration},\cite{c22},\cite{c23},\cite{c24}, a larger area can be searched, and more victims can be located and rescued efficiently by using a heterogeneous aerial-ground team. 
In the agriculture field \cite{c25},\cite{c26},\cite{c27}, the team with heterogeneous aerial-ground robots has been proved a powerful tool to accomplish supervising and farming tasks; 
In \cite{c28}, \cite{c29}, \cite{c30}, heterogeneous aerial-ground teams have been used to reduce task complexity by assigning sub-tasks to different robots. For example, while a UAV locates victims who require medical treatment, the UAV cannot assist the victim directly. Therefore, the nearest UGV is requested to deliver medical supplies to complete the task. 

\begin{figure}[!t]
  \centering
 \includegraphics [width=0.95 \columnwidth ]{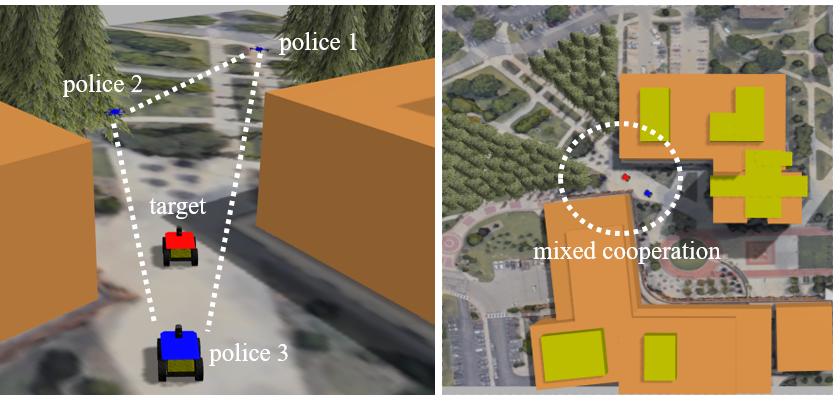}
  \caption{With the help of Proficiency Awareness mixed cooperation, police 1 invited capable team members (police 2$\And$3) to work on a tracking task.}
  \label{fig:overview}
  \vspace{-1em}
\end{figure}

Complex and dynamic conditions pose challenges for establishing efficient aerial-ground robot teams. First, robots are generally designed for specific conditions and tasks, while the real-world tasks are dynamic and complex so it is more difficult to deploy and scale \cite{c36},\cite{c37},\cite{c38}. Secondly, real-world factors such as motor degradation and sensor failure make the actions of faulty robots unpredictable and therefore decrease the cooperation effectiveness \cite{c39},\cite{c40},\cite{c41}.
Thirdly, the diversity of capabilities of individual robots makes it challenging to design an algorithm to integrate all the capabilities on one aerial-ground robot team \cite{c6},\cite{c35}. However, ignoring the heterogeneous robot capabilities will limit the potential of mixed aerial-ground teaming in real-world applications. 
Therefore, there is an urgent need to develop a methodology to maximize the advantage and simplify the controlling of mixed aerial and ground robot teams.

To address this challenge, this paper proposed a proficiency aware multi-agent deep reinforcement learning method (Mix-RL) to formulate robot proficiency. As shown in Fig. \ref{fig:overview}, using the Mix-RL, an aerial-ground robot team can flexibly deploy strategies to complete tasks in dynamic environments based on the awareness of the robot teammates' capabilities. The main contributions of this research are as follows: 
\begin{itemize}
    \item[(1)]
    A proficiency aware multi-agent actor-critic method has been proposed to exploit the potential of a mixed aerial-ground team by optimizing the team configuration.
    \item [(2)]
    A reinforcement learning-based mixed aerial-ground robot cooperation framework has been developed to dynamically design strategies for tasks based on robot capabilities, task requirements, and environment conditions.
\end{itemize}

\section{Related Work}

Prior work investigated mixed teaming of heterogeneous robots. To complete searching and rescuing tasks,\cite{c8},\cite{c30},\cite{c16}, systems for UAVs and UGVs collaboration were designed to allocate specific tasks to different team members. However, their studies are unsuitable for real-time applications in unexpected situations. These methods have a burdensome hierarchical controlling structure that requires transmitting information to a central decision maker to generate strategies. In this paper, each team member understands the proficiency of individual members and directly request help from others. We provided a more efficient cooperation mechanism using a decentralized control strategy with proficiency awareness. A decentralized mixed cooperation scheme was investigated \cite{c10},\cite{c31} where the drone and the human operator provided guidance to UGVs for navigating among obstacles. \cite{c32} investigated the synergistic integration of aerial and ground vehicles, which showed complementary capabilities. However, the study did not explore the in-process changes of proficiency, making it difficult to allocate the tasks to suitable teammates according to real-time conditions. Our paper utilized the actor-critic learning algorithm, which allows each member to criticize the behaviors of its teammates and learn to self-correct in a real-time manner.

There are applications that leverage the flexibility of multi-robot teams to perform tasks in human-hostile environments \cite{dynamicWildfireTracking}, \cite{c11}. The multi-robot tasks in these applications were dynamic and cooperative. However, the applications only involved uniformed robot types and were limited by prior knowledge of the working map. Some researches also investigated mixed robot team for navigating, tracking or searching tasks with centralized controlling architecture \cite{c10}, \cite{c16}. However, the robot control in these research was either limited in inter-teammate awareness or heavily depended on human attention. Given that, the proficiency-aware mixed cooperative method in our paper presented a more efficient and flexible way for multi-robot systems to work in a complex and dynamic environment.

\section{Robot Proficiency Aware Mixed Cooperation and Competition}
In this section, a proficiency-aware multi-agent deep reinforcement learning algorithm was proposed to support heterogeneous robot cooperation, which is derived from \cite{maddpg}. The algorithm utilizes a framework of centralized training with decentralized execution, which eases the training process in a non-stationary environment; this model also integrated with a proficiency-aware mechanism, which exploits the advantage of the resilient capabilities of robot to facilitate robot teaming in different environments.

\subsection{Multi-Agent Deep Reinforcement Learning}

\textbf{Preliminaries.} Considering a multi-robot cooperation and competition problem in which the robots have shared and conflicting goals. The problem with $N$ robots typically consists of a sequential set of states $S$ modelling the positions and the conditions of the robots, a set of robot actions $A_1,A_2,...,A_N$ constricting all the safe movements according to the specific physical features of robots, and a set of observations $O_1,O_2,...,O_N$ for each robot. 
Each robot decides how to take action according to the policy $\boldsymbol{\pi}_{\theta_i}(a_i|o_i)$, where policy parameters are denoted by $\theta = \{\theta_1,...,\theta_N\}$. By adopting the policy, robots reach new states described by the state transition function $T: S \times A_1 \times A_2 \times ... \times A_N \mapsto S$. Upon reaching the new state, each robot receives a reward $r_i: S \times A_i \mapsto \mathbb{R}$. Restricted by perceiving ability of robot and real world limitations, each robot obtains a distinctive local observation $o_i: S \mapsto O_i$. This paper considers a case of two multi-robot teams with different objectives and receiving varied even adversarial rewards. Each robot aims at maximizing its own expected aggregate objective,
\begin{equation}
    J(\theta_i) = \mathbb{E}_{s \sim S,a_i \sim \boldsymbol{\pi}_i}[R_i]
    \label{eq:RobotObjective}
\end{equation}
where discounted reward $R_i = \sum^{TIME}_{t=0} \gamma^{t}r_{i,t}$ and discounted factor $\gamma \in [0,1]$.

\textbf{Multi-Robot Actor-Critic Learning.} Robots are given the actions $a = (a_1,...,a_N)$ and the observations $s = (O_1,O_2,...,O_N)$ of the robots of both teams to perceive the world. Once the robots are equipped with extra information, the world can be seen as fully known and be treated as stationary despite the changing of policies. Then the gradient of (\ref{eq:RobotObjective}) with deterministic policies can be written as:
\begin{equation}
    \begin{aligned}
    {}
    & \nabla_{\theta_i} J(\theta_i) = \\
    & \mathbb{E}_{s \sim S^{\boldsymbol{\mu}},a_i \sim \boldsymbol{\pi_i}}[\nabla_{\theta_i} \log \boldsymbol{\pi_i} (a_i|o_i)Q^{\boldsymbol{\pi}}_i(s,a)|_{a_i=\boldsymbol{\mu}(o_i)}]
    \end{aligned}
    \label{eq:gradientfunc}
\end{equation}
where $Q_i^{\boldsymbol{\pi}}(s,r_1,...,r_N)$ is the extra information for the robots at training time, which includes the observations of all the robots and selected state information $s = (o_1,...,o_N,s_1,...,s_N)$. The team's awareness to exploit the most capable robots in cooperation can be cultivated through updating the policy along (\ref{eq:gradientfunc}) to minimize the regression loss:
\begin{equation}
\begin{aligned}
    {}
    & \mathcal{L}(\theta_i) = \mathbb{E}_{s,a,r,s'}[(Q_i^{\boldsymbol{\mu}}(s,a)-y)^2] \\
    & \text{where\quad} y = r_i + \gamma Q_i^{\boldsymbol{\mu}'}(s',a')|_{a'_j=\boldsymbol{\mu}'_j(o_j)}
    \label{eq:lossfunc}
\end{aligned}
\end{equation}

\subsection{Proficiency Awareness Modeling for Mixed Teaming}
Selecting robots with proper capabilities to perform tasks in a complex environment assures task success. This paper defines the environment as $\mathbb{E}_{\mathbb{T}}=\{e_1, e_2, ...\}$, where $\mathbb{T}$ is the assigned task, and $e$ is the environmental condition such as lawn or forest. Considering the differences in capabilities of robots, this paper defines the capabilities of robot $i$ as $\mathbb{A}_{i}: \{\mathbb{A}_{i}^{e}=(\alpha_1^e, \alpha_2^e, ...), e \in \mathbb{E}_{T}$\}, where $\alpha_i$ is the capability of robot $i$ such as velocity or observation range. $\mathbb{A}_{i}$ means robot i has different capabilities such as velocity and observation range for different environmental conditions such as lawn or forest.
The proficiency $f_p(\mathbb{E}, \mathbb{T}, \mathbb{A})$ is defined as:
\begin{equation}
\begin{aligned}
    &f_p(\mathbb{E}, \mathbb{T}, \mathbb{A})=\prod \limits f_{\eta}(e,\mathbb{A}_{i}^{e}), e \in \mathbb{E}_{T}, i \in N
    \label{eq:ProficiencyFunc}
\end{aligned}
\end{equation}
if $\forall e \in \mathbb{E}_{T}$ and $\forall i \in N, f_p(\mathbb{E}, \mathbb{T}, \mathbb{A}) > \xi$, the mixed teaming is with proficiency.
N is the number of robots in the team. $f_{\eta}$ is the measurement of efficiency of a team of robots with different capabilities $\mathbb{A}_{i}^{e}$ performing a task in different environment conditions, such as $e_1$ and $e_2$. $\xi$ is a predefined efficiency threshold. If one UAV and one UGV - the capabilities are denoted as $\mathbb{A}_1, \mathbb{A}_2$ respectively - are selected to perform the task together, then $f_p(\mathbb{E}, \mathbb{T}, \mathbb{A})=f_{\eta}(e_1,\mathbb{A}_{1}^{e_1},\mathbb{A}_{2}^{e_1})\times f_{\eta}(e_2,\mathbb{A}_{1}^{e_2},\mathbb{A}_{2}^{e_2})>\xi$, the mixed teaming is with proficiency.

Considering the situation with multiple robots ($R_o$), this paper defines the awareness of proficiency as the robot set $\sum r_o^{\ast}$ that the robot who has a greater capability to the specific environment has a larger probability to be chosen to perform the task. The awareness of proficiency is denoted as:
\begin{equation}
\begin{aligned}
    &P(\sum r_{o}^{\ast}) \\
    &\forall \sum r_{o}\in R_{o}:f_p(\mathbb{E}, \mathbb{T}, \mathbb{A}_{\sum r_o^{\ast}})\geq f_p(\mathbb{E}, \mathbb{T}, \mathbb{A}_{\sum r_o})
\end{aligned}
\end{equation}
Overall, the increase in $P(\sum r_{o}^{\ast})$ means the robot team has a better awareness of proficiency and performs the assigned task with more efficiency based on this awareness.

The overall reward in the training process of Reinforcement Learning, $Reward_i$, includes two separate reward functions: objective reward function and proficiency reward functions. 
\begin{equation}
    Reward_i = R_{i,objective}+R_{i,position}
\end{equation}

Robots’ awareness of proficiency is introduced by the proficiency reward function, $R_{i,position}$, which is modelled by robot motion constraints and observations.
\begin{equation}
R_{i,position} = \beta Velocity_{i}(s) + \gamma Observation_{i}(s)
\label{eq:posRwd}
\end{equation}
$\beta$ and $\gamma$ are the balance weights. The proficiency reward function indicates robot has varied capability when it is in different regions of the environment. The differences in capability between robots, including mobility and real-time observation range, determine their different $R_{i,position}$ for every state. For example, a robot will have a negative $Velocity_{i}(s)$ while it is in an incapable state $s$. On the other hand, objective reward function $R_{i,objective}$ describes the distance between robots and its target position measures how well a robot fulfills its objective.

\section{Evaluation}
The effectiveness of mixed cooperation and competition for robot teaming could be reflected through effective behaviors of mixed UAV-UGV teams in a task under dynamic environments. In this section, we evaluated mixed UAV-UGV behaviors with a surveillance task under the simulation environment of Kent State University Student Center, University Library and Risman Plaza. The environment was simulated with CRAImrs simulation platform.

The performance of the Mix-RL method was compared with the original multi-agent actor-critic algorithm to show the improvement in the task reward, success rate of task, appropriate involvement of individual robots.

\subsection{Experiment Environment Settings} \label{settings}

\begin{figure}[!t]
\begin{minipage}{.95\columnwidth}
    \centering
	\begin{minipage}{0.81\columnwidth}
		\begin{figure}[H]
			\includegraphics [width=1.0 \columnwidth ]{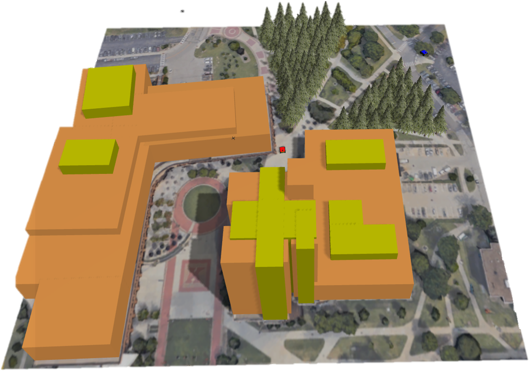}
			
		\end{figure}
	\end{minipage}
	\begin{minipage}{0.17\columnwidth}
		\begin{figure}[H]
		\centering
            \begin{subfigure}[b]{\columnwidth}
                \centering
                \includegraphics [width=\columnwidth ]{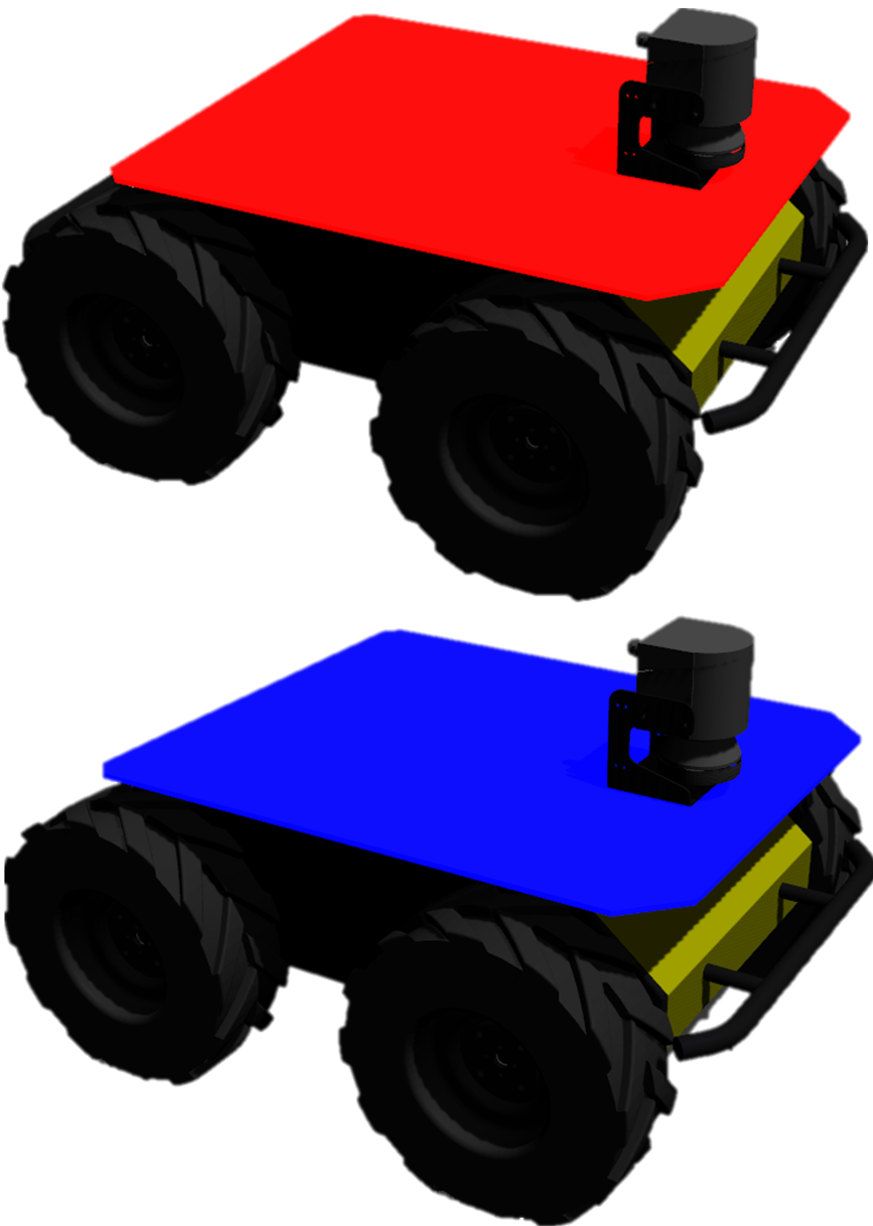}
                \caption{Huskys}
                \label{fig:huskys}
            \end{subfigure}

            \begin{subfigure}[b]{\columnwidth}
                \centering
                \includegraphics [width=\columnwidth ]{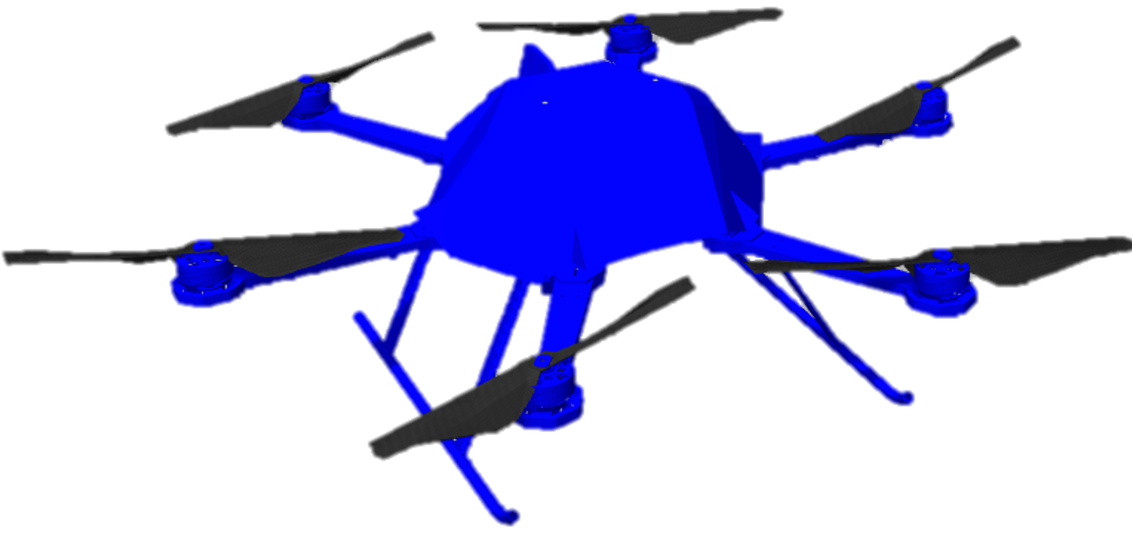}
                \caption{Firefly}
                \label{fig:firefly}
            \end{subfigure}
            \begin{subfigure}[b]{\columnwidth}
                \centering
                \includegraphics [width=\columnwidth ]{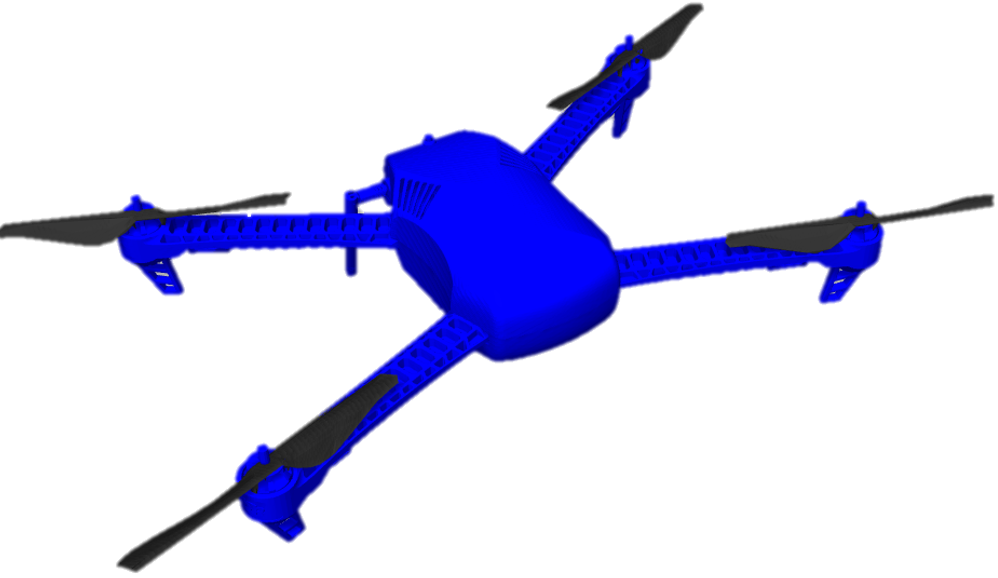}
                \caption{Iris}
                \label{fig:iris}
            \end{subfigure}
		\end{figure}
	\end{minipage}
	
\end{minipage}
\caption{The Simulation Environment and robot settings \cite{Furrer2016}.}
\label{fig:illustration}
\end{figure}

This paper used three different kinds of vehicles in the experiment. \textit{Firefly} has a redundant 6-rotor propulsion system that is very robust and guarantees a stable flight, as shown in Figure \ref{fig:firefly}, \cite{c3}. In our experiment, \textit{Firefly} has a wide range of perception but moves slower than \textit{Iris}. \textit{Iris} is a quad-copter, shown in Figure \ref{fig:iris}. It is not as stable as \textit{Firefly}, but it can fly faster \cite{c4}. \textit{Iris} moves fast but has a relatively small range of perception. \textit{Husky} is a ground robotic platform, shown in Figure \ref{fig:huskys}. Its max speed is 1.0 m/s.

\begin{figure}[ht]
    \begin{subfigure}{.45\columnwidth}
    \centering
    {Criminal Tasks\\}
    \vspace{5pt}
    
    \includegraphics[width=\columnwidth]{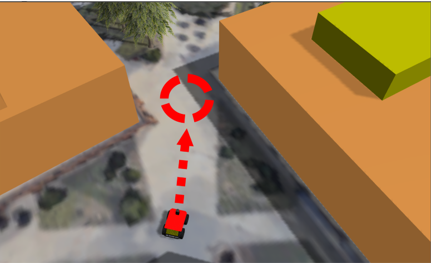}
    \caption{Reaching Target Locations}
    \end{subfigure}
    \begin{subfigure}{.45\columnwidth}
    \centering
    {Police Tasks\\}
    \vspace{5pt}
    \includegraphics[width=\columnwidth]{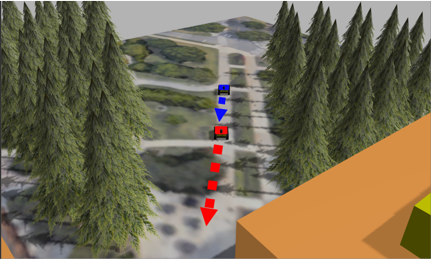}
    \caption{Pursuing \quad Criminal (Red)}
    \end{subfigure}

    \begin{subfigure}{.45\columnwidth}
    \centering
    \includegraphics[width=\columnwidth]{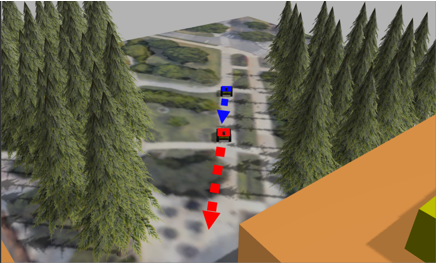}
    \caption{Escaping from Police (Blue)}
    \end{subfigure}
    \begin{subfigure}{.45\columnwidth}
    \centering
    \includegraphics[width=\columnwidth, height=0.607\columnwidth]{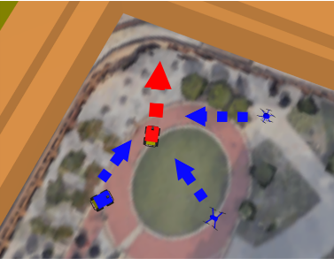}
    \caption{Surrounding Criminal (Red)}
    \end{subfigure}
\caption{Different tasks of Police and Criminal}
\label{fig:tasks}
\end{figure}

The vehicles were divided into two teams: the police and criminal. 
The criminal team only consists of UGV. Its goal is to get close to SoIs and avoid getting caught, as depicted in Figure \ref{fig:tasks}. 
The police team includes three different types of robots, \textit{Firefly}, \textit{Iris} and  \textit{Husky}. The police team tries to capture the criminal robot, and when the police team surrounds the criminal robot, the capture is successful. 
In the experiment, the success of a police task is defined as a successful capture of a criminal robot before it reaches any SoI. The minimum number of police set to 2, and the maximum distance required for a capture set to 1.0 meter between the criminal robot and the two police robots.

This paper built a 3D world based on Kent State University campus map in CRAImrs, shown in Figure \ref{fig:illustration}. Based on real situations, UAV and UGV models both have limited observable ranges and speed in the simulation. The max speed, max acceleration, and perceiving range radius of each robot in the experiment are shown in Table \ref{tab:robotpara}.

\begin{table}[t]

\centering
\small
\caption{Parameters of UAV and UGV}
\resizebox{\columnwidth}{!}{%
    \begin{tabular}{c c c c}
    \hline\noalign{\smallskip}
    Type     & V$_{max}$ ($m/s$) & Acceleration ($m/s^{2}$) & Radius (m) \\
    \noalign{\smallskip}\hline\noalign{\smallskip}
    \textit{Firefly} & 5.0 & 1.0   & 30  \\
    \textit{Iris}    & 7.0 & 2.0   & 30   \\
     \textit{Husky}   & 1.0 & 0.1 & 15   \\
    \noalign{\smallskip}\hline\noalign{\smallskip}
    \end{tabular}%
}
\label{tab:robotpara}
\end{table}

This paper validated the effectiveness of Mix-RL in the aspect: robot function adaptive mixed cooperation.

\noindent\textbf{Robot Function Adaptive Mixed Cooperation.} The hypothesis is that the Mix-RL could adapt different robots in mixed cooperation.
In this experiment, we only considered the observation range and ignore other observation abilities such as imaging quality \cite{c17}. 
One robot can sense all the position information of the robots that are in the range of observation.
This paper also had the setting that UAVs have a larger observation range compared with UGVs.
In the expected adaptive mixed cooperation, the police team should consist of two fastest UAVs in the open area to effectively perform the task.

\subsection{Multi-robot Cooperation Analysis}

To evaluate the improvement of multi-robot cooperation, three different metrics were introduced in the experiment: captured engagement rate, task success rate and police-criminal's achieved reward.
The task success rate and reward metrics both show that the Mix-RL method improved robot cooperation. With proficiency awareness, the two most capable robots were tended to cooperate. In the condition of open areas, two fastest UAVs were assigned to complete the pursuing tasks.

\begin{figure}[ht]
  \centering
  \includegraphics [width=0.95 \columnwidth ]{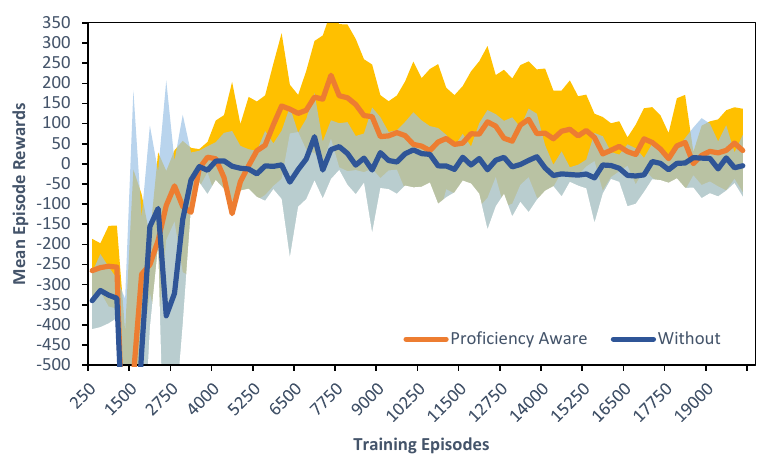}
  \caption{Learning curves of mixed robot cooperation with and without proficiency awareness. We trained for 20,000 episodes with both methods. The plots report the mean episode rewards during the training and the error bars are calculated with 95\% confidence interval over 5 runs.}
  \label{fig:learning curve with CI}
  \vspace{-1em}
\end{figure}

\textbf{Proficiency Awareness.} The performance of multi-robot with proficiency awareness was demonstrated by if the deployment of robots adapts to the dynamic environment given the differences in robot's mobility, flexibility and observation range. During the testing phase, situations were observed in open area where police UAVs chase the criminal robot. Because: 1) UGV have a lower average speed compared with UAVs; 2) There are plenty of police UAVs for one ground target. Therefore, the high involvement of police UAV in the task contributes to a successful capture. In contrast, without using the proficiency-aware method, results show that the robot team failed to cooperate, which means the lack of participation of capable robots in the task is the main reason for multi-robot cooperation failures.

\textbf{Cooperation Effectiveness.} 
The mix-RL model can improve the proficiency of multi-robot cooperation. Here we adopted task success rates to measure the effectiveness and reliability of multi-robot cooperation. The algorithm with proficiency awareness maintained a high task success rate. The task success rate of robot team with-awareness in Task Plaza was 89.60\%, while the task success rate of robot team without-awareness was 55.20\%. Besides, as shown in Figure \ref{fig:learning curve with CI}, the mean reward of training episodes with proficiency-aware setting was $71.29$. However, The mean reward of training episodes without proficiency-aware setting was $-16.33$.
Therefore, the mix-RL model can improve the efficiency of cooperative and competitive aerial-ground robot teaming in different scenarios

\section{CONCLUSION}

This paper proposed a proficiency-aware actor-critic method (mix-RL), which improved the success rate and efficiency of cooperative and competitive aerial-ground robot teaming. We deployed a heterogeneous UAV-UGV team to perform tasks in ground target tracking scenarios. The effectiveness of mix-RL in improving task success rates is presented to validate the feasibility of deploying this method for guiding the flexible teaming of mixed aerial-ground robot in dynamic environments. In the robot function-adaptive situation, with proficiency-aware setting, the robot team could generate appropriate teaming engagement to leverage the different capabilities of individual robots. In the future, one important research can be the scalability of the method with a larger number of heterogeneous robots and targets.


\begin{thebibliography}{99}

\bibitem{c3} AscTec Firefly http://www.asctec.de/en/uav-uas-drones-rpas-roav/asctec-firefly/
\bibitem{c4} Iris - The Ready to Fly UAV Quadcopter  http://www.arducopter.co.uk/iris-quadcopter-uav.html
\bibitem{c6} Parker L.E. (1996) Multi-Robot Team Design for Real-World Applications. In: Asama H., Fukuda T., Arai T., Endo I. (eds) Distributed Autonomous Robotic Systems 2. Springer, Tokyo
\bibitem{c8} Connie Phan and Hugh H.T. Liu A Cooperative UAV/UGV Platform for Wildfire Detection and Fighting 


\bibitem{c10} El Houssein Chouaib Harik, François Guérin, Frédéric Guinand1,Jean-François Brethé, Hervé Pelvillain. UAV-UGV Cooperation For Objects Transportation In An Industrial Area. 978-1-4799-7800-7/15/31.00 2015 IEEE.

\bibitem{dynamicWildfireTracking} Pham, Huy X., et al. "A distributed control framework for a team of unmanned aerial vehicles for dynamic wildfire tracking." 2017 IEEE/RSJ International Conference on Intelligent Robots and Systems (IROS). IEEE, 2017.

\bibitem{c11} Jingjing Gu, Tao Su, Qiuhong Wang, Xiaojiang Du, Mohsen Guizani. Multiple Moving Targets Surveillance Based on a Cooperative Network for Multi-UAV. IEEE Communications Magazine, pages 82-89, 2018.

\bibitem{c16} Owen,  M.,  Yu,  H.  McLain,  T., Beard.  R.  2010. Moving  Ground  Target  Tracking  in  Urban Terrain  Using  Air/Ground  Vehicles,  IEEE  GLOBECOM  Workshops  (GC  Wkshps).  Miami,  FL-USA, December 6-10, pp 1816–1820  

\bibitem{c17} Gazebo Tutorial\\ http://gazebosim.org/tutorials?tut=ros\_gzplugins\#Camera

\bibitem{resilient_disaster_site_exploration} Gregory, J. M., Brookshaw, I., Fink, J., and Gupta, S. K. (2017, October). An investigation of goal assignment for a heterogeneous robotic team to enable resilient disaster-site exploration. In 2017 IEEE International Symposium on Safety, Security and Rescue Robotics (SSRR) (pp. 133-140). IEEE.

\bibitem{maddpg} Lowe, R., Wu, Y., Tamar, A., Harb, J., Abbeel, O. P., and Mordatch, I. (2017). Multi-agent actor-critic for mixed cooperative-competitive environments. In Advances in Neural Information Processing Systems (pp. 6379-6390).

\bibitem{Furrer2016} Furrer, F., Burri, M., Achtelik, M., and Siegwart, R. (2016). RotorS-A modular Gazebo MAV simulator framework. In Robot Operating System (ROS) (pp. 595-625). Springer, Cham.

\bibitem{c22} A. Gawel, R. Dubé, H. Surmann, J. Nieto, R. Siegwart and C. Cadena, "3D registration of aerial and ground robots for disaster response: An evaluation of features, descriptors, and transformation estimation," 2017 IEEE International Symposium on Safety, Security and Rescue Robotics (SSRR), Shanghai, 2017, pp. 27-34, doi: 10.1109/SSRR.2017.8088136.
2016.

\bibitem{c23} Ivan Maza, Antidio Viguria, Anibal Ollero ”Networked aerial-ground robot system with
distributed task allocation for disaster
management”Safety, Security and Rescue Robotics, 2006

\bibitem{c24} Nathan Michael, Shaojie Shen, Kartik Mohta, Yash Mulgaonkar, and Vijay Kumar ”Collaborative Mapping of an Earthquake-Damaged
Building via Ground and Aerial Robots” Journal of Field Robotics 29(5), 832–841 (2012)

\bibitem{c25} P. Menendez-Aponte, C. Garcia, D. Freese, S. Defterli and Y. Xu, "Software and Hardware Architectures in Cooperative Aerial and Ground Robots for Agricultural Disease Detection," 2016 International Conference on Collaboration Technologies and Systems (CTS), Orlando, FL, 2016, pp. 354-358, doi: 10.1109/CTS.2016.0070.

\bibitem{c26} P. Tokekar, J. V. Hook, D. Mulla and V. Isler, "Sensor Planning for a Symbiotic UAV and UGV System for Precision Agriculture," in IEEE Transactions on Robotics, vol. 32, no. 6, pp. 1498-1511, Dec. 2016, doi: 10.1109/TRO.2016.2603528.

\bibitem{c27} A. Vasudevan, D. A. Kumar and N. S. Bhuvaneswari, "Precision farming using unmanned aerial and ground vehicles," 2016 IEEE Technological Innovations in ICT for Agriculture and Rural Development (TIAR), Chennai, 2016, pp. 146-150, doi: 10.1109/TIAR.2016.7801229.

\bibitem{c28} M. Langerwisch, T. Wittmann, S. Thamke, T. Remmersmann, A. Tiderko and B. Wagner, "Heterogeneous teams of unmanned ground and aerial robots for reconnaissance and surveillance - A field experiment," 2013 IEEE International Symposium on Safety, Security, and Rescue Robotics (SSRR), Linkoping, 2013, pp. 1-6, doi: 10.1109/SSRR.2013.6719320.

\bibitem{c29}B. Arbanas et al., "Aerial-ground robotic system for autonomous delivery tasks," 2016 IEEE International Conference on Robotics and Automation (ICRA), Stockholm, 2016, pp. 5463-5468, doi: 10.1109/ICRA.2016.7487759.

\bibitem{c30}Cai Luo, A. P. Espinosa, D. Pranantha and A. De Gloria, "Multi-robot search and rescue team," 2011 IEEE International Symposium on Safety, Security, and Rescue Robotics, Kyoto, 2011, pp. 296-301, doi: 10.1109/SSRR.2011.6106746.

\bibitem{c31}El Houssein Chouaib Harik, François Guérin, Frédéric Guinand, Jean-François Brethé, Hervé Pelvillain. A Decentralized Interactive Architecture for Aerial and Ground Mobile Robots Cooperation.
ICCAR 2015. 2015 IEEE International Conference on Control, Automation and Robotics, May 2015,
Singapour, Singapore. pp.37-43, ff10.1109/ICCAR.2015.7165998ff. ffhal-01145928

\bibitem{c32}Grocholsky, Ben; Keller, James; Kumar, R. Vijay; and Pappas, George J., "Cooperative Air and Ground Surveillance" (2006).Departmental Papers (MEAM).Paper 74 

\bibitem{c35}Lynne E. Parker. (2001) Evaluating success in autonomous multi-robot teams: experiences from ALLIANCE architecture implementations. Journal of Experimental \& Theoretical Artificial Intelligence 13:2, pages 95-98.

\bibitem{c36}G. Antonelli, F. Arrichiello and S. Chiaverini, "Stability analysis for the Null-Space-based Behavioral control for multi-robot systems," 2008 47th IEEE Conference on Decision and Control, Cancun, 2008, pp. 2463-2468, doi: 10.1109/CDC.2008.4738697.

\bibitem{c37}L. Vig and J. A. Adams, "Multi-robot coalition formation," in IEEE Transactions on Robotics, vol. 22, no. 4, pp. 637-649, Aug. 2006, doi: 10.1109/TRO.2006.878948.

\bibitem{c38}C. Jones and M. J. Mataric, "Adaptive division of labor in large-scale minimalist multi-robot systems," Proceedings 2003 IEEE/RSJ International Conference on Intelligent Robots and Systems (IROS 2003) (Cat. No.03CH37453), Las Vegas, NV, USA, 2003, pp. 1969-1974 vol.2, doi: 10.1109/IROS.2003.1248936.

\bibitem{c39}K. LeBlanc and A. Saffiotti, "Cooperative anchoring in heterogeneous multi-robot systems," 2008 IEEE International Conference on Robotics and Automation, Pasadena, CA, 2008, pp. 3308-3314, doi: 10.1109/ROBOT.2008.4543715.

\bibitem{c40}S. C. Botelho and R. Alami, "M+: a scheme for multi-robot cooperation through negotiated task allocation and achievement," Proceedings 1999 IEEE International Conference on Robotics and Automation (Cat. No.99CH36288C), Detroit, MI, USA, 1999, pp. 1234-1239 vol.2, doi: 10.1109/ROBOT.1999.772530.

\bibitem{c41}{Dasgupta, P. (2011). Multi-robot task allocation for performing cooperative foraging tasks in an initially unknown environment. In L. Jain, E. Aidman, E. Aidman, \& C. Abeynayake (Eds.), Innovations in Defence Support Systems -2: Socio-Technical Systems (pp. 5-20). (Studies in Computational Intelligence; Vol. 338). https://doi.org/10.1007/978-3-642-17764-42}


\end{thebibliography}
\end{document}